\documentclass[11pt,a4paper]{article}
\usepackage[hyperref]{acl2021}
\usepackage{times}
\usepackage{latexsym}
\usepackage{graphicx}

\usepackage{paralist}
\usepackage{microtype}
\aclfinalcopy
\usepackage{multirow}

\usepackage{color}

\title{Sattiy at SemEval-2021 Task 9: An Ensemble Solution for Statement Verification and Evidence Finding with Tables}

\author{Xiaoyi Ruan, Meizhi Jin, Jian Ma, Haiqin Yang$^{\S}$,  Lianxin Jiang, \\ {\bf Yang Mo} and {\bf Mengyuan Zhou}\\
  Ping An Life Insurance Co., Ltd. \\
  Shenzhen, Guangdong province, China \\
  {\small\{RUANXIAOYI687, EX-JINMEIZHI001, MAJIAN446, JIANGLIANXIN769, MOYANG853\}@pingan.com.cn}\\
  $^{\S}$ {\small the corresponding author, email: {hqyang@ieee.org}}
}

\date{}

\begin{document}
\maketitle
\begin{abstract}
Question answering from semi-structured tables can be seen as a semantic parsing task and is significant and practical for pushing the boundary of natural language understanding.  Existing research mainly focuses on understanding contents from unstructured evidence, e.g., news, natural language sentences, and documents.  The task of verification from structured evidence, such as tables, charts, and databases, is still less explored.  This paper describes sattiy team’s system in SemEval-2021 task 9: Statement Verification and Evidence Finding with Tables (SEM-TAB-FACT).  This competition aims to verify statements and to find evidence from tables for scientific articles and to promote the proper interpretation of the surrounding article.  In this paper, we exploited ensemble models of pre-trained language models over tables, TaPas and TaBERT, for Task A and adjust the result based on some rules extracted for Task B.  Finally, in the leaderboard, we attain the F1 scores of 0.8496 and 0.7732 in Task A for the 2-way and 3-way evaluation, respectively, and the F1 score of 0.4856 in Task B. 
\end{abstract}

\section{Introduction}
{
Semantic parsing is one of the most important tasks in natural language processing.  It not only needs to understand the meaning of natural language statements, but also needs to map them to meaningful executable queries, such as logical forms, SQL queries, and Python code~\cite{pan2019recent,lei2020re,zhu2021retrieving}.  Question answering from semi-structured tables is usually seen as a semantic parsing task~\citep{DBLP:conf/acl/PasupatL15}, where questions are translated into logical forms that can be executed against the table to retrieve the correct denotation~\cite{DBLP:journals/corr/abs-1709-00103}.

Practically, it is significant in natural language understanding to verify whether a textual hypothesis is entailed or refuted by evidence~\cite{2008A,D1978The}.  The verification problem has been extensively studied in different natural language tasks, such as natural language inference (NLI)~\cite{DBLP:journals/corr/BowmanAPM15}, claim verification~\citep{DBLP:journals/corr/abs-1809-01479}, recognizing of textual entailment (RTE)~\citep{2005The}, and multi-model language reasoning (NLVR/NLVR2) \citep{DBLP:journals/corr/abs-1811-00491}.  However, existing research mainly focuses on verifying hypothesis from unstructured evidence, e.g., news, natural language sentences and documents.  Research of verification under structured evidence, such as tables, charts, and databases, is still in the exploratory stage. 
}

This year, SemEval-2021 Task 9: Statement Verification and Evidence Finding with Tables (SEM-TAB-FACT), aims to verify statements and find evidence from tables in scientific articles~\cite{wang2021semeval}.  It is an important task targeting at promoting proper interpretation of the surrounding article.  

The competition tries to explore table understanding from two tasks:
\begin{compactitem}[--]
\item Task A - Table Statement Support: The task aims to determine whether a statement is fully supported, refuted, or unknown to a given table.
\item Task B - Relevant Cell Selection: given a statement and a table, the task is to determine which cells in the table provide evidence for supporting or refuting the statement.  \if 0 More specifically, a table cell is evidence for the statement if it helps support or refute a part of the statement:
\begin{compactitem}
\item {\em Relevant}: the cell must be included.
\item {\em Ambiguous}: the cell is allowed to be either included or not included.
\item {\em Irrelevant}: the cell must not be included.
\end{compactitem}
\fi 
%each cell and each statement, if the cell is within the mini- mum set of cells needed to provide evidence for the statement (”relevant”) or not (”irrelevant”). statement.
\end{compactitem}

{
The competition contains the following challenges:
\begin{compactitem}[--]
\item In task A, there is no training data for the ``Unknown" category and the number of tables is small in the training set.
\item The lexical expression of the table may be different from that in the statement.
\item The table structure is complex and diverse.  A table may contain missing values while the the supporting evidence may be resided in cells across several rows or columns.  % cells to support the statement may  in several rows or  several columns. 
\item It is difficult to understand the statements.  For example, some statements express totally different semantics meaning with only one different word.  This difficulty makes it even harder to find the evidence cells from tables.
%\item More seriously, 
\end{compactitem}
}

\par To overcome these challenges, we incorporate several key technologies in our implementation:
\begin{compactitem}[--]
\item developing a systematic way to generate data from the ``Unknown" category;
\item including additional data corpus to enrich the training data;
%\item  Align table and statement content.
\item exploiting existing state-of-the-art pre-trained language models over tables, TaBERT~\cite{DBLP:conf/acl/YinNYR20} and  TaPas~\cite{2020arXiv200508314Y}, and ensembling them into a powerful one;
\item aligning contents in tables and statements while constructing manual rules for tackling Task B.
\end{compactitem}
The test shows that our implementation can increase the performance according and finally, in the leadboard, we attain the F1 scores of 0.8496 and 0.7732 in Task A for the 2-way and 3-way evaluation, respectively, and the F1 score of 0.4856 in Task B.

The rest of this paper is organized as follows: In Sec.~2, we briefly depict related work to our implementation.  In Sec.~3, we detail our proposed system.  In Sec.~4, we present the experimental setup and analyze the results.  Finally, we conclude our work in Sec.~5.
\if 0
\begin{table*}
\centering
\caption{Lable distribution in all tables.\label{citation-guide}}
\begin{tabular}{|c|c|c|c|c|c|c|}
\hline
\textbf{Type} & \textbf{\# tables}  & \textbf{Entailed} & \textbf{Refuted} & \textbf{Unknown} & \textbf{\# T. tokens} & \textbf{\# S. tokens}\\
\hline
Task 9 & 981 & 2,818 & 1,688 & 0 & \\ % (63\%) (37\%) 
TabFact & 16,573 & 63,962 & 54,313 & 0 &\\ % (54\%)  (46\%)  
Augmented & 17,554 & 66,780 & 56,001 & 61,436 & \\ % (36.3\%) (30.4\%) (33.3\%)
\hline
\end{tabular}
\end{table*}

\begin{table*}
\centering
\begin{tabular}{|c|c|c|c|c|c|c|}
\hline
      & \multicolumn{2}{c|}{max} & \multicolumn{2}{c|}{min} & \multicolumn{2}{c|}{mean} \\ \hline
      & row      & statement     & row      & statement     & row      & statement      \\ \hline
train & 302      & 88            & 1        & 3             & 10       & 11             \\ \hline
dev   & 115      & 53            & 2        & 4             & 11       & 13             \\ \hline
test  & 69       & 82            & 2        & 4             & 11       & 12             \\ \hline
\end{tabular}
\caption{Token counts distribution in all tables.\label{citation-guide}}
\end{table*}
\fi 

\begin{table*}[htp]
\centering
\begin{tabular}{|c|c|c|c|c|c|}
\hline
 & & \multirow{2}{*}{Tables} & Label Distribution     &  Tokens in Statements & Tokens in Tables \\ \cline{4-6} 
     &             &                   & Entailed/Refuted/Unknown & Max./Min./Avg.            & Max./Min./Avg.      \\ \hline
%\parbox[t]{2mm}{\multirow{3}{*}{\rotatebox[origin=c]{90}{Train}}} 
\multirow{3}{*}{Train} & Task 9             & 981                     & 2,818/1,688/0           & 88/3/11                & 302/1/10         \\ \cline{2-6}
& Tabfact           & 16,573                  & 63,962/54,313/0         & 57/4/14                & 127/5/13         \\ \cline{2-6}
& Augm.        & 17,554                  & 66,780/56,001/61,436    & 88/3/12                & 302/1/11         \\ \hline
Dev. & $-$ & 52 & 250/213/93 & 53/4/13 & 115/2/11 \\\hline
Test & $-$ & 52 & $-$ & 82/4/12 & 69/2/11 \\\hline
\end{tabular}
\caption{Data statistics.\label{tab:data}}
\end{table*}

\begin{figure*}[ht]
    \centering
    \includegraphics[width=\textwidth]{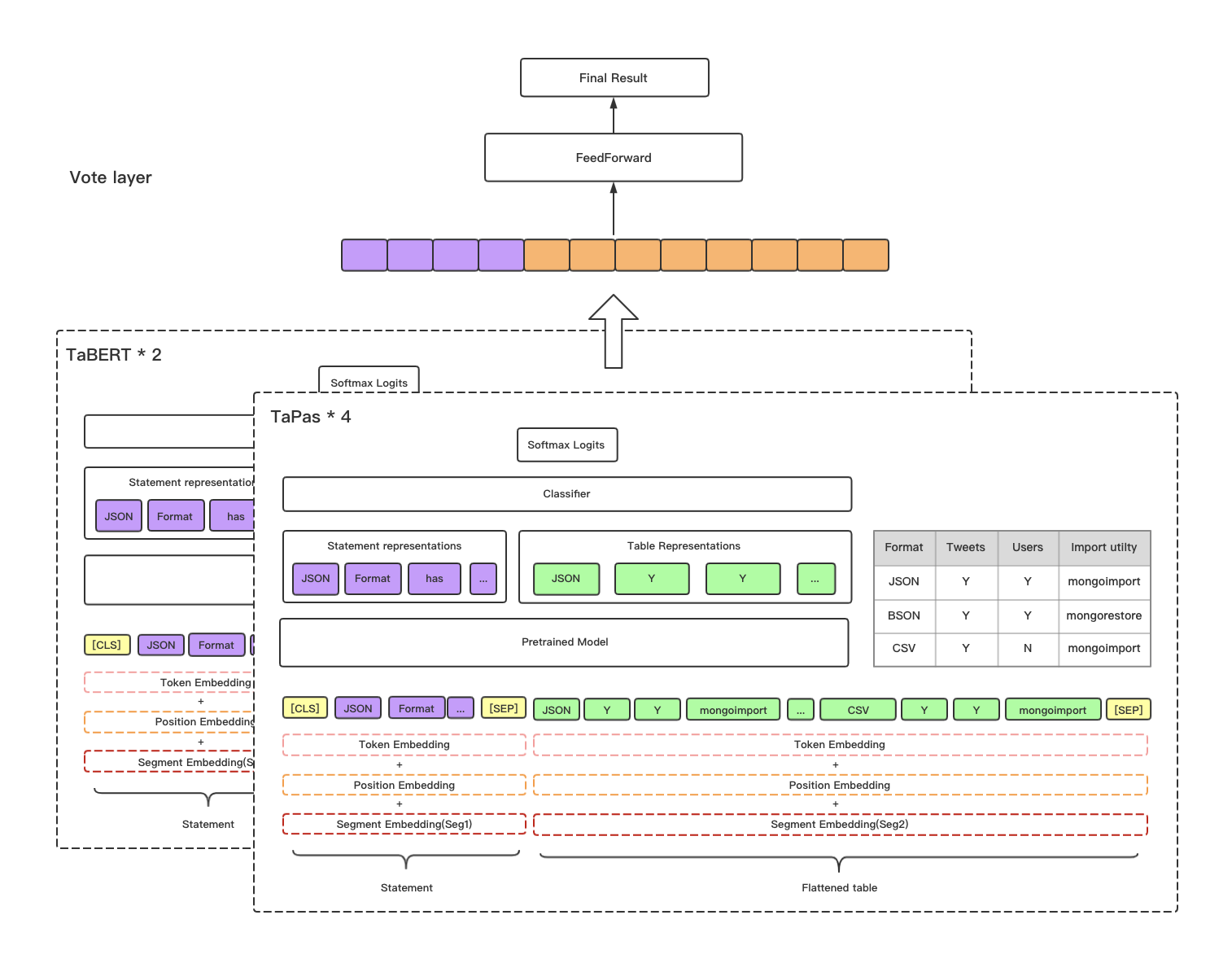}
    \caption{Ensemble architecture}
    \label{fig.main}
\end{figure*}
\section{Related Work}

Recently, pre-trained language models (PLMs), e.g., BERT~\citep{DBLP:conf/naacl/DevlinCLT19}, XLNET~\citep{DBLP:journals/corr/abs-1906-08237}, and  RoBERTa~\citep{DBLP:journals/corr/abs-1907-11692}, have witnessed the burgeoning of promoting various downstream NLP tasks, such as reading comprehension, named entity recognition and text classification~\cite{li2020molweni,lei2021have}.   However, the current pretrained language models are basically trained on the general text.  They are not fit for some tasks, e.g., Text-to-SQL, Table-to-Text, which need to encode the structured data, because the data in the structured table also needs to be encoded at the same time.  Directly applying the existing PLMs may face the problem of inconsistency between the encoded text from the table and the pretrained text.

TaBERT\citep{DBLP:conf/acl/YinNYR20} is a newly proposed pretrained model built on top of BERT and jointly learns contextual representations for utterances and the structured schema of database (DB) tables.  This model views the verification task completely as an NLI problem by linearizing a table as a premise sentence and applies PLMs to encode both the table and statements into distributed representation for classification.  This model excels at linguistic reasoning like paraphrasing and inference but lacks symbolic reasoning skills.  Intuitively, encoding more table contents, e.g., type information and content snapshots, relevant to the input utterance could potentially help answer questions that involve reasoning over information across multiple rows in the table because they can provide more hints about the meaning of a column. TaPas~\citep{DBLP:conf/acl/HerzigNMPE20} is another  newly proposed pretrained question answering model over tables implemented on BERT to avoid generating the logical forms.  The model can fine-tune on semantic parsing datasets, only using weak  supervision, with an end-to-end differentiable recipe. 

Another stream of work on evidence finding with table is the rule-based approaches.  Most evidence cells can be extracted by rules.  For example, if a row head or column head appears in the statement, we infer this row or col support this statement. Although rule-based approaches suffer from the low recall issue, they exhibit high precision and can be applied to adjust the result for ensemble.

\section{System Overview}
We elaborate the task and present our system in the following.
\subsection{Data Description and Tasks}
In Task A, the original dataset is a set of XML files, where each XML file represents a table and contains multiple sections:
\begin{compactitem}[--]
\item {\em document} section represents the whole document;
\item {\em table} section determines the unique ID of the document;
\item {\em caption} section is a brief description of the table;
\item {\em legend} section is a detailed description of the table;
\item {\em multiple row} sections describe the contents of each row of the table; and 
\item {\em statements} section provides several factual statements.
\end{compactitem}
This task aims to determine if a statement is entailed or refuted by the given table, or whether, as is in some cases, this cannot be determined from the table.  The competition also provides two kinds of evaluations for the task:
{
\begin{compactitem}
\item 3-way F1 score evaluation: a standard precision/recall evaluation (3-way) is computed to evaluates whether each table is correctly classified into one of the three types in  \{Entailed, Refuted, Unknown\}.  It is to test whether the classification algorithm understands cases where there is insufficient information to make a determination.
\item 2-way F1 score evaluation: the F1 score is computed to evaluate the performance when the statements with the ``unknown" ground truth label are removed.  The metric will also penalize misclassifying Refuted/Entailed statement as unknown. 
\end{compactitem}
In the evaluation, the score for all statements in each table is first averaged and then averaged across all tables to get the final F1 score. }

In Task B, the raw dataset is a subset of task A, where unknown statements are excluded.  The goal is to determine for each cell and each statement, if the {\em cell is within the minimum set of cells} needed to provide evidence for the statement ``relevant" or ``irrelevant".  For some statements, there may be multiple minimal sets of cells that can be used to determine statement entailment or refusal.  In such cases, the ground truth will contain all of the  versions.  The evaluation will calculate the recall and precision for each cell, with ``relevant" cells as the positive category.  The evaluation is conducted similarly as that in Task A.

%Similar to Task A, the score will be averaged over all statements in each table first, before proceeding to average across all tables. 

%we need to determine which cells in the table can support the current statement.  \red{In the evaluation, we first average the score for all statements in each table and average them afterwards as the final F1 score.}

%For score averaging, we first average the score for all statements in each table, which is then averaged across all tables for the final F1 score.

\subsection{Data Augmentation}
There are mainly two critical issues in Task A.  First, the number of the tables is small.  We then include more external data, the TabFact dataset~\citep{DBLP:conf/iclr/ChenWCZWLZW20} to improve the generalization of our proposed system.  Second and more critically, ``unknown" statements do not exist in the training set but may appear in the test set.  To allow our system to output the ``unknown" category, we construct additional ``unknown" statements to enrich the training set.  More specifically, we randomly select some statements from other tables and assign them to the ``unknown" category for the current table.  In order to keep balance on the labels, the number of selected statements from other tables is set to half of the statements in the current table.  Details about the data statistics can be referred to Table~\ref{tab:data}.

\subsection{Model Ensemble for Task A}
Figure 1 outlines the overall structure of our system, which is an ensemble of two main pretrained models on table-based data, TaBERT and TaPas, or two variants of TaBERT and four variants of TaPas.   {It is worth noting that the input of all models are the same.  That is, given a statement and a table, the input is started with the sentence token, [CLS], followed by the sequence of the tokens in the statement, the segmentation token ([SEP]), and the sequence of the tokens in the flattened table.  All the tokens in the statement and the table are extracted by wordpiece as in BERT and related NLP tasks~\cite{DBLP:conf/naacl/DevlinCLT19,DBLP:journals/corr/abs-1909-09292,conf/ijcai/PHED21,conf/ijcnn/EDM21,conf/ijcnn/RefBERT21}.}  {The flattened table means that we borrow the implementation in TaBERT by only extracting the most likely content snapshot as detailed in Sec.~3.4.  The obtained tokens' embeddings are then fed into six strong baselines, i.e., two variants of TaBERT and four variants of TaPas, to attain the classification scores for the corresponding labels.  The classification scores are then concatenated and fed into a vote layer, i.e., a fully-connected network, to yield the final prediction.  %, i.e., three scores for each model for Task A and two scores for each model for Task B.  
}

\begin{table*}[!htb]
\centering
\begin{tabular}{|c|c|c|c|}
\hline \textbf{Models} & \textbf{Original data} & \textbf{+TabFact}& \textbf{+Augm.} \\ \hline
{TaBERT\_1}& 0.7446/0.6580& 0.7634/0.6837& 0.8003/0.7502\\% {TaBERT\_Large\_(K=1)}
{TaBERT\_3}& 0.7689/0.6792& \textbf{0.7952}/0.7008& \textbf{0.8241/0.7653}\\%{TaBERT\_Large\_(K=3)}
{TaPas\_TFIMLR} & {0.7502/0.6637}& {0.7859/0.6799}& {0.8102/0.7649}\\ % TaPas tabfact inter masklm large reset
{TaPas\_WSIMLR}& {0.7498/0.6522}& {0.7852/0.7005}& {0.8024/0.7577}\\ % TaPas wikisql sqa inter masklm large reset
{TaPas\_IMLR} & {0.7538/0.6358}& {0.7799/0.6890}& {0.7908/0.7396} \\ % TaPas inter masklm large reset
{TaPas\_WSMLR}& \textbf{0.7695/0.6875}& {0.7904/\textbf{0.7058}}& {0.8156/0.7609}\\ % TaPas\_wikisql\_sqa\_masklm\_large\_reset
\hline
\end{tabular}
\caption{\label{tab:rs_task_A}Comparison of strong baselines in Task A for 2-way and 3-way evaluation.}
\end{table*}

\begin{table}[!htb]
\centering
\begin{tabular}{|@{~}c@{~}|@{~}c@{~}|@{~}c@{~}|@{~}c@{~}|}
\hline \textbf{Models} & \textbf{+TabFact}& \textbf{+Augm.} & \textbf{+Rule}  \\ \hline
{TaBERT\_1} & 0.4025& 0.4159& 0.4605\\% {TaBERT\_Large\_(K=1)}
{TaBERT\_3} & 0.4158& {0.4305}& {0.4685}\\% {TaBERT\_Large\_(K=3)}
%{TaPas\_tabfact\_inter\_masklm\_large\_reset}
{TaPas\_TFIMLR} & {0.4253}& {0.4299}& {0.4597}\\
%{TaPas\_wikisql\_sqa\_inter\_masklm\_large\_reset}
{TaPas\_WSIMLR} & {0.4199}& {0.4208}& {0.4682}\\
TaPas\_IMLR & {0.4006}& {0.4102}& {0.4467}\\
%{TaPas\_inter\_masklm\_large\_reset}
%{TaPas\_wikisql\_sqa\_masklm\_large\_reset}
{TaPas\_WSMLR} & \textbf{0.4258}& \textbf{0.4386}& \textbf{0.4708}\\
\hline
\end{tabular}
\caption{\label{tab:rs_task_B}Comparison of strong baselines in Task B.}
\end{table}

\subsection{Content Snapshot}
In order to pin point the important rows and avoid excessively encode input from the table, we borrow the idea of content snapshot in TaBERT~\cite{DBLP:conf/acl/YinNYR20} to encode only a few rows that are most relevant to the statement.  We create the content snapshot of $K$ rows based on the following simple strategy.  First, we count the number of rows of each table and find their median, say $R$.  If the number of rows in the current table is less than or equal to $R$, then $K$ is set to the total number of rows in the current table and the content snapshot is the entire content of the current table.  If the number of rows in the current table is greater than $R$, we set $K=R$ and select the top-$K$ row with the highest overlap rate between the statement and each row of $n$-grams as the candidate rows.

\subsection{Rule Construction for Task B}
For Task B, we apply the same model trained in Task A to find whether the current table supports the statement.  If yes, we label all cells as entailed.  Otherwise, we first align the word expression in tables and statements while building the corresponding rules to adjust the model prediction.  That is, we change uppercase to lowercase and transform all abbreviations into the full name in statements, cells, col heads and row heads.  We also conduct stemming on all words.  For example, ``definition" and ``defined" is transformed to ``define".  After that, we collect all words in a statement into a word bag and determine the supporting relation based on the following rules: 1) If a word in the word bag appears in a row head, we then infer that cells in the whole column supports the statement; 2)If a word appears in the first column of the table, we then infer that cells in the whole row supports the statement; 3) If a word appears in both a row head and a cell in the first column of a table, we then infer that the cell corresponding to the row and column supports the statement; 4) If a word appears in a cell, we then infer that this cell supports the statement.
\if 0
\begin{compactitem}[--]
\item If a word in the word bag appears in a row head, we then infer that cells in the whole column supports the statement;
\item If a word appears in the first column of the table, we then infer that cells in the whole row supports the statement;
\item If a word appears in both a row head and a cell in the first column of a table, we then infer that the cell corresponding to the row and column supports the statement;
\item If a word appears in a cell, we then infer that this cell supports the statement.
\end{compactitem}
\fi 

\section{Experiments}
In the following, we present the strong baselines and the results with analysis.  

We have tried different combinations of TaBERT and TaPas pre-trained models and choose the following 6 best baselines: 1) TaBERT\_1: the pre-trained TaBERT with $K=1$; 2) TaBERT\_3: the pre-trained TaBERT with $K=3$; 3) TaPas\_TFIMLR: the pre-trained large TaPas downloaded from tapas\_tabfact\_inter\_masklm\_large\_reset.zip; 4) TaPas\_WSIMLR: the pre-trained large TaPas downloaded from tapas\_wikisql\_sqa\_inter\_masklm\_large\_reset.zip; 5) TaPas\_IMLR: the pre-trained large TaPas downloaded from tapas\_inter\_masklm\_large\_reset.zip; 6) TaPas\_WSMLR: the pre-trained large TaPas downloaded from  tapas\_wikisql\_sqa\_masklm\_large\_reset.zip.
\if 0
\begin{compactitem}
\item TaBERT\_1: the pre-trained TaBERT with $K=1$; 
\item TaBERT\_3: the pre-trained TaBERT with $K=3$;
\item TaPas\_TFIMLR: the pre-trained large TaPas downloaded from tapas\_tabfact\_inter\_masklm\_large\_reset.zip
\item TaPas\_WSIMLR: the pre-trained large TaPas downloaded from tapas\_wikisql\_sqa\_inter\_masklm\_large\_reset.zip; 
\item TaPas\_IMLR: the pre-trained large TaPas downloaded from tapas\_inter\_masklm\_large\_reset.zip;
\item TaPas\_WSMLR: the pre-trained large TaPas downloaded from  tapas\_wikisql\_sqa\_masklm\_large\_reset.zip
\end{compactitem}
\fi 
Our proposed system is funetuned on the above models for the original training data, the original data with the TabFact data, and the augmentation data.  We also tune the hyper parameters to fit a better result in the local test dataset.

Table~\ref{tab:rs_task_A} reports the evaluation results of Task A on the development set when funetuning the above six strong baselines on different training data.  The results show that the TaPas\_WSMLR attains the best performance on the original data.  The best performance is further improved from 0.7695 to 0.7952 for 2-way evaluation and from 0.6875 to 0.7058 for 3-way evaluation, respectively, by including the TabFact data.  The performance is further improved to 0.8241 for 2-way evaluation and 0.7653 for 3-way evaluation, respectively, by adding the augmentation data.  Finally, we apply the voting mechanism to ensemble the results and achieve the F1 scores of 0.8496 and 0.7732 on the test set, respectively.

Table~\ref{tab:rs_task_B} reports the results of Task B on the development set when funetuning the above six strong baselines on different training data.  The results show that the TaPas\_WSMLR attains the best performance among all six strong baselines and the perform increases from 0.4258 after adding the TabFact data, to 0.4386 after adding the augmentation data, and to 0.4708, additional 7.3\% improvement after adding the manual rules.  We conjecture that TaPas\_WSMLR can provide more complementary information for solving the task.   Finally, we ensemble the results by the voting mechanism and achieve the F1 score of 0.4856 on the test set.  

In sum, results in Table~\ref{tab:rs_task_A} and Table~\ref{tab:rs_task_B} confirm the effectiveness of our proposed system by including more training data and the manual rules.  

\section{Conclusion}
In this paper, we present the implementation of our ensemble system to solve the problem of SemEval 2021 Task 9.  To include more training data and resolve the issue of lacking data from the ``Unknown" category in the training set, we include external corpus, the TabFact dataset, and specially construct the augmented data for the ``Unknown" category.  Content snapshot is also applied to reduce the encoding effort.  Six pre-trained language models over tables are funetuned on the TabFact dataset and the augmented data with content snapshot tables to evaluate the corresponding performance.  An ensemble mechanism is applied to get the final result.  Moreover, data alignment and manual rule determination are applied to solve Task B.  Finally, our system attains the F1 score of 0.8496 and 0.7732 in Task A for 2-way and 3-way evaluation, respectively, while getting the F1 score of 0.4856 in Task B. 

\bibliographystyle{acl_natbib}
\bibliography{acl2021}
\end{document}